# 3D CONCEPTUAL DESIGN USING DEEP LEARNING

**Zhangsihao Yang**
zhangsiy@andrew.cmu.edu

**Haoliang Jiang**
haolianj@andrew.cmu.edu

**Lan Zou**
lzou1@andrew.cmu.edu

## ABSTRACT

This article proposes a data-driven methodology to achieve a fast design support, in order to generate or develop novel designs covering multiple object categories. This methodology implements two state-of-the-art Variational Autoencoder dealing with 3D model data. Our methodology constructs a self-defined loss function. The loss function, containing the outputs of certain layers in the autoencoder, obtains combination of different latent features from different 3D model categories.

Additionally, this article provide detail explanation to utilize the Princeton ModelNet40 database, a comprehensive clean collection of 3D CAD models for objects. After convert the original 3D mesh file to voxel and point cloud data type, we enable to feed our autoencoder with data of the same size of dimension. The novelty of this work is to leverage the power of deep learning methods as an efficient latent feature extractor to explore unknown designing areas. Through this project, we expect the output can show a clear and smooth interpretation of model from different categories to develop a fast design support to generate novel shapes. This final report will explore 1) the theoretical ideas, 2) the progresses to implement Variantional Autoencoder to attain implicit features from input shapes, 3) the results of output shapes during training in selected domains of both 3D voxel data and 3D point cloud data, and 4) our conclusion and future work to achieve the more outstanding goal.

## 1 INTRODUCTION

Three-dimensional(3D) representations of real-life objects has been an essential application in computer vision, robotics, medicine, virtual reality and mechanical engineering. Due to the increase of interest in tapping into latent information of 3D models in both academic and industrial areas, generative neural network of 3D objects have achieve impressive improvement in the last several years. Recent research achievement on generative deep learning neural network concepts includes autoencoder (Rohit Girdhar, 2016), Variational Autoencoder (VAE) (Alreza Makhzani, 2016) and generative adversarial model (Ian Goodfellow, 2014). These models have led great progresses in natural language processing (Junyoung Chung, 2015), translate machine and image generation (Danilo Jimenez Rezende, 2016). With the capability of extract features from data, these models are also widely used in generating embedding vectors (Danilo Jimenez Rezende, 2016). In studies of 3D shapes, researchers utilized these state-of-the-art generative neural networks to take latent feature vectors as representation of 3D voxelized shapes and 3D point cloud shapes. Those researches show promising results during experiments. To support the 3D object research community, several large 3D CAD model database came out, such as notably ModelNet (Zhirong Wu, 2014) and ShapeNet (Angel X. Chang, 2015).

Although the generative deep learning neural networks are able to extract general latent features from training database, as well as to apply variable changes to the output shape, the output 3D models are still restricted to multiple certain design areas: it cannot compute or conclude the result when important part of the input data is in lack. As a result, we exploited recent advances in generative model of 3D voxel shapes (Shikun Liu, 2017), 3D point cloud shapes (Panos Achlioptas, 2018) and a series of research work in fast image and video style transfer proposed initially by leon A. Gatys et al. (leon A. Gatys, 2015). In this paper, the authors came up with the concept of style and the content of two input images and used a loss function to combine the feature extracted from both images. After that, convolutional neural network was used in real-time style transfer (Justin Johnson, 2016) and videos style transfer(Xun Huang, 2017). Motivated by those works, we propose a multi-model architecture with a patten-balanced loss function to extract implicit information from multiple input objects , and set the implicit information as the constrains on the output of our deep learning neural network. By using our methodology, we are able to combine the latent features to our output 3D model and generate totally new design areas.

Briefly, the contributions of this project are shown as follows:

- To propose a novel multi-model deep learning neural network and define a multi-pattern loss function to learn implicit features from input objects from the same category using 3D point cloud and objects from different categories using 3D voxel data.

- To show that our model is capable of realizing shape transfer which will be helpful to break current design domains and generate brand new novel designs.

- To demonstrate the potentiality of implementing generative neural network to develop fast 3D model design support based the methodology this project proposed

The general pipeline of our project is:

- Data processing: convert mesh file into 3D voxel data and 3D point clouds data
- Building a Variational Shape Learner building and training to verify its performance
- Building a Point Net Autoencoder building and training to verify its performance
- Implementing VSL and Point Net Autoencoder into our systems
- Experiment: applying different layer outputs in loss function; taking multiple categories input pattern, like furniture and transportation

## 2 BACKGROUND

3D Object Representation. In this project, the authors are tapping into voxel data and point cloud as 3D object representation. Voxel can represent a value on a 3D regular grid. Voxel data are capable to represent a simple 3D structures with both of empty and filled cubes, frequently used to display volumetric objects. Point Cloud data is a type of 3D space representation using a set of 3D data points. It is generally produced by 3D scanners to sample a large number of points from the surface of a space objects. Point Clouds are widely used as a detailed 3D object representation, also used as 3D CAD models for designing and manufacturing.

Autoencoders. Autoencoders are generative deep neural network architectures, designed to extract features from the input and reproduce the output. Usually, they contain a narrow bottleneck layer as the low dimensional representation or an embedding code for the input data. There are two parts of Autoencoders: encoder and decoder. The encoder is to down sample the input to the latent code, the decoder to expand the latent code to reproduce the input data. Variational Autoencoder (VAE) models are autoencoders taking advantage of the distribution of latent variables. In VAE, variational approach is used to learn latent representation. As a result, VAE contains additional loss component and specific training algorithm.

Neural Style Transfer. Neural style transfer (NST) is a trending topic since 2016. It can be interpreted as the process of using Convolutional Neural Network (CNN) to embed certain styles into image or video contents. It is widely studied in both of academic literatures and industrial applications, and received increasing attention from industries currently. A variety of approaches, not limited to CNN, is used to solve or explore new aspects. Nevertheless, 3D style transfer is still an unexplored topic which could be very useful in 3D graphic area.

Figure 1: Architecture of Neural Network

## 3 NETWORK ARCHITECTURE

The network architecture we are using is inspired by the fast style transform. In this architecture, there are one auto-encoder and three loss providers, with each of them has different functions.

The auto-encoder is used to transform input data to the desired output style and is called transform machine. The other three auto-encoders are utilized to provide loss for the loss function, and each of them could be called separately: content loss provider, style loss provider and prediction loss provider.

The assumption in the network architecture is that certain layers' outputs could provide information about a 3D model's content and that certain layers outputs could offer information about a 3D model's style. The 2D content and style refers to the content and the style of a image. For example, when offer one image of New York City at daylight condition and one image of Pittsburgh at nightlight condition, the output preserve the content of New York City and the style of Pittsburgh. This means you will get a image of New York City at moonlight condition. In this article, however, the content and style are refer to input 1 and input 2, so we could change the mixture level of content and style with higher flexibility by converting content and style as needed, to get the scene of what is content and style based on different scenario.

Take the style transform from an airplane to a car as an example. First, the airplane is put into the auto-encoder, the output is called transformed airplane. Then put the transformed airplane, the original airplane and the original car into three different loss providers to generate loss functions for back-propagation in the network. One thing to be noticed is that: the loss provider does not do any back-propagation. So the weights in the loss provider are fixed during training, while the only changing weights during training period is the weights of the style transform auto-encoders. Regarding more detail information about the reason why we fix part of the network, the fast style transform article (Xun Huang, 2017) could be a related reference to interested readers.

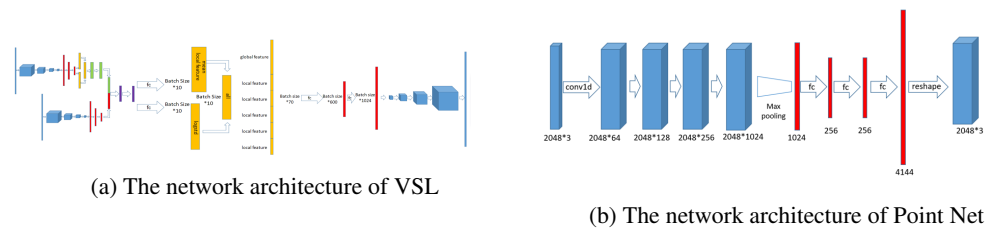

(a) The network architecture of VSL

(b) The network architecture of Point Net

Figure 2: The network architecture

The figure above shows the overall architecture of couple different neural networks in the system. To be more specific, our network system could experiment on different state of art auto-encoder network architecture (figure 2). The networks in used are VSL and Point Net auto-encoder version (Panos Achlioptas, 2017). VSL (Variational Shape Learner) is a network with voxel input and the voxel formatted data output. Point Net auto-encoder is with point cloud data as input and the output.

## 4 DATA

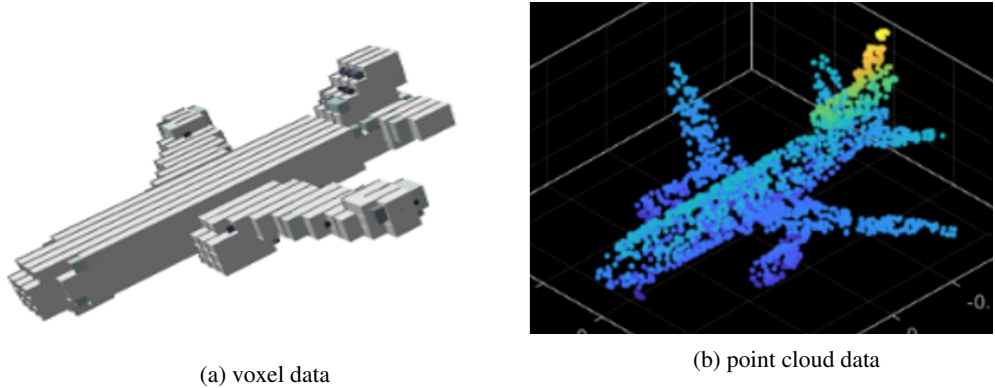

(a) voxel data

(b) point cloud data

Figure 3: Visualization of voxel data and point cloud data

The data in the network architecture is voxel data and point cloud data. The visualization of the data is shown in figure 3. The voxel data is either 0 or 1, with data shape in (30, 30, 30). The point cloud data's shape is in (number of points, points dimension). There are 3 dimension for each point in this case. When using Point Net network, there are 2048 certain number of points.

## 5 EXPERIMENT & RESULTS

In this task, firstly per-trained the original auto-encoder for 3000 epochs only based on the cars and the airplanes dataset. Then use this network as the auto-encoder and as the loss providers.

We use the outputs of layers of content loss provider as the target content, and outputs of layers of style loss provider as the target style. The predicted target content and target style are extracted from the corresponding outputs of transformed object's loss provider's layer. Then compute MSE loss on these outputs to do back-propagation. These results are shown in the figure 4, 5, and 6.

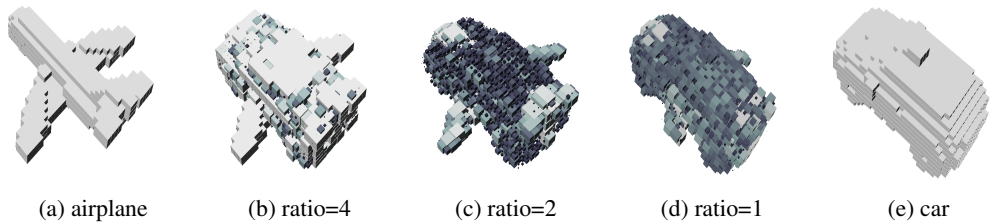

(a) airplane    (b) ratio=4    (c) ratio=2    (d) ratio=1    (e) car

Figure 4: Shape transform between airplane and car

The results of using car and airplane are shown in figure 4. When we change the ratio of the loss of airplane and car, the output of the auto-encoder will change as well. Indicated from left to right, as the ratio of airplane becoming lower, the output is approximating to a car. As the evidence to avoid simple merge of a car and a airplane, the wings of the airplane are becoming smaller as the ratio of the airplane decreased.

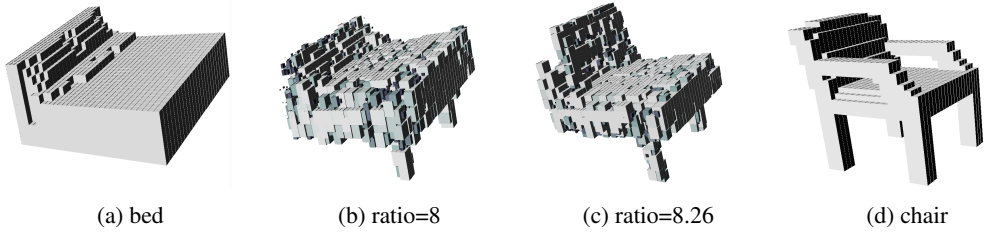

(a) bed     (b) ratio=8     (c) ratio=8.26     (d) chair

Figure 5: Shape Transform between Airplane and Car

Additionally, the results of transform between chairs and tables are shown in figure 5. It seems that, at certain ratio, the chair and the bed become one recliner which makes this transforming work more promising.

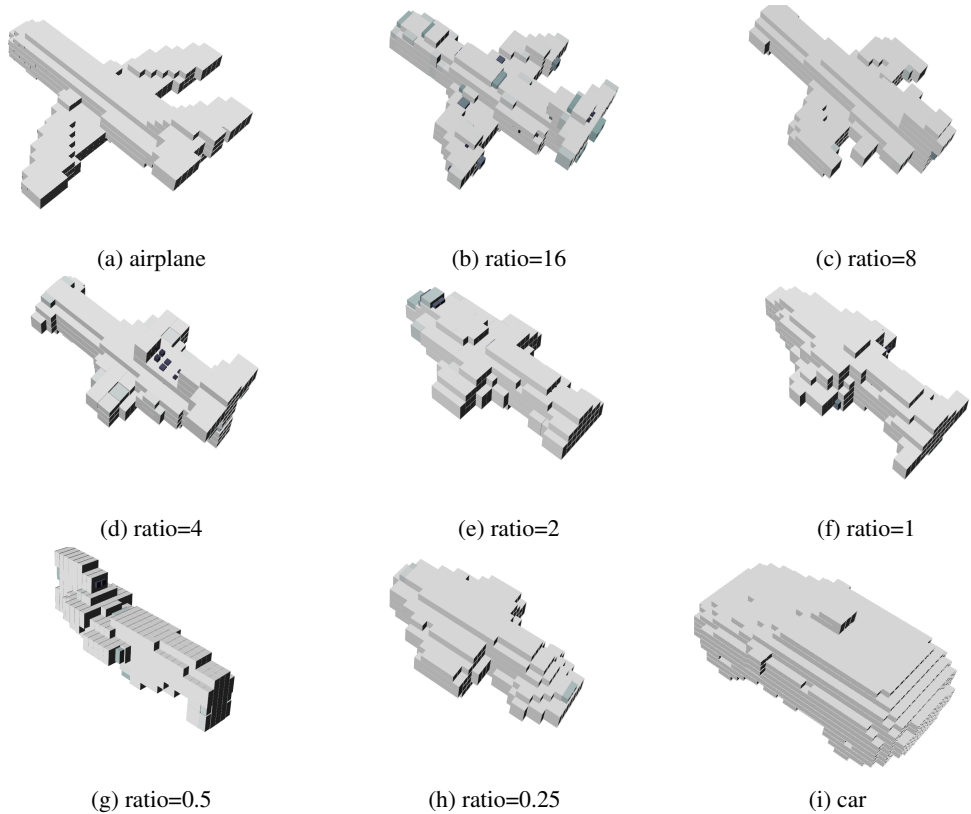

(a) airplane     (b) ratio=16     (c) ratio=8

(d) ratio=4     (e) ratio=2     (f) ratio=1

(g) ratio=0.5     (h) ratio=0.25     (i) car

Figure 6: Transform between Airplane and Car When Using Global Features and Local Features

Based on the VSL paper (Shikun Liu, 2017), where local features and global features are named in similar situation as ours, we decided to use local features for the style of images and global features as the content of images. The reason that the figure 6 looks out of imagination is because that the output is becoming something like space ships.

When using Point Net auto-encoder, the results are shown in figure 7, 8, 9, and 10. The content layers using is first 1d convolution output. And the style layer used is the bottleneck feature of the network.

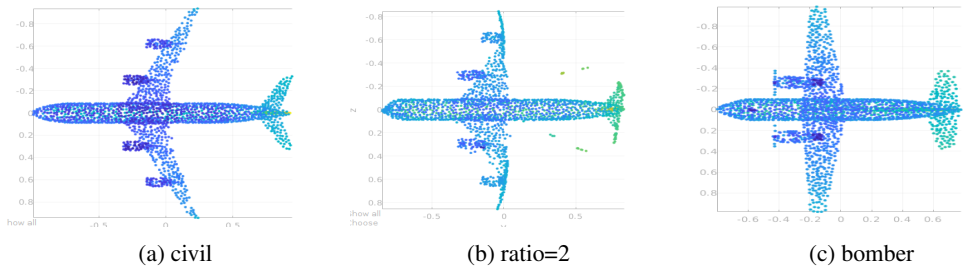

(a) civil     (b) ratio=2     (c) bomber

Figure 7: Transform result between a civil and a bomber

In figure 7, the left airplane (a) is a civil airplane and the right one (b) is a bomber airplane. When combining both of them together, the wings are combing two wings together in the shape of a civil airplane and mimic the style of the bomber straight wings. The number of engine number is same as the civil airplane, this means that the network truly learn the content of civil airplane. The wings and the horizontal stabilizer mimic the style of the bomber airplane.

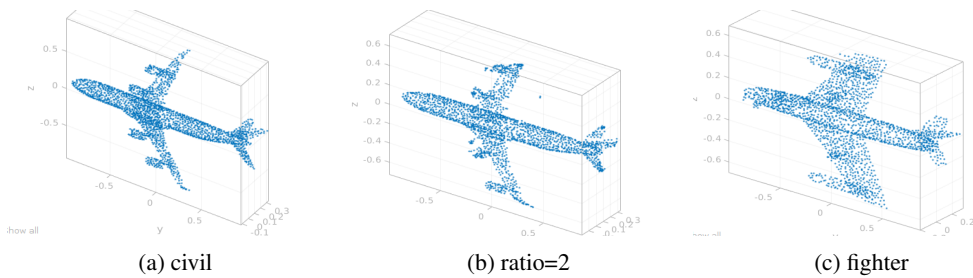

(a) civil     (b) ratio=2     (c) fighter

Figure 8: Transform result between a civil and a fighter

In figure 8, combining a civil airplane (a) and a fighter airplane (c), the generated airplane has learnt how a airplane could change. The shape of the airplane has changed its length and shape. The fuselage length mimic the civil which is regard as the input of a content loss provider. And the fuselage's shape mimic the fighter's shape. Additionally, the wings' shape and horizontal stabilizer changed its' shape to a combination of both the wings of the civil aircraft's and fighter's.

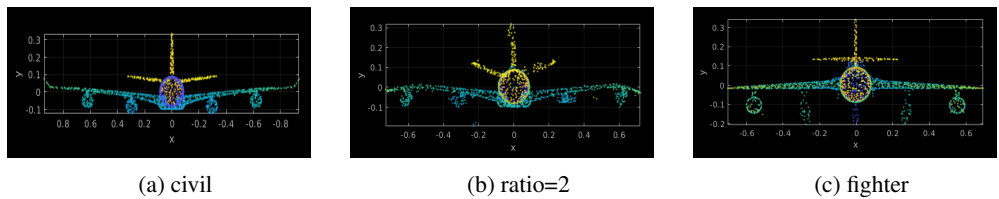

(a) civil     (b) ratio=2     (c) fighter

Figure 9: front view of the result between a civil and a fighter

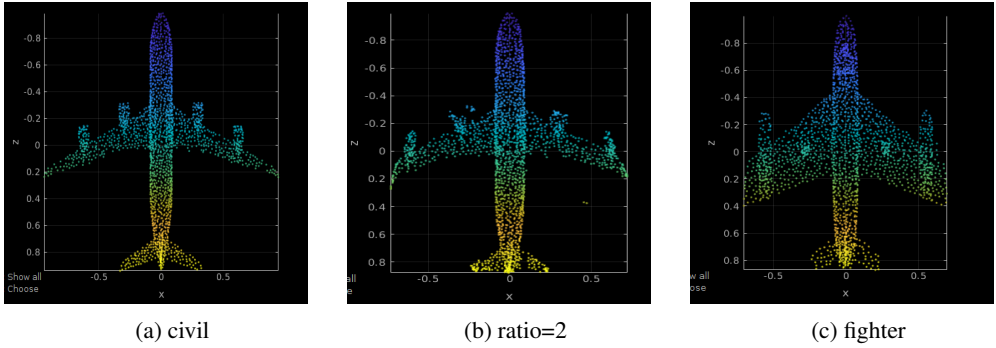

| (a) civil | (b) ratio=2 | (c) fighter |

Figure 10: top view of the result between a civil and a fighter

In figure 9 and figure 10, the front view and the top view provide more detailed information about how the wings and the horizontal stabilizers change.

## 6 CONCLUSION

This article proposes a data-driven methodology and implements two different 3D shape representations and VAEs into the methodology. Though the increasing interest in NST, 3D shape transfer is still an unexplored area for future research and experiments. 3D shape transfer could be very useful both in academic areas and industrial applications. The project tapped into voxel data, Point Cloud data, Point Net and VSL to see the performance. The experiments implement via voxel data, our system is able to generate novel shapes which contain latent features of the inputs. By varying the ratio in loss function and in the layers where the loss function extracts output, the output of the system will change locally and globally. The experiments set point cloud as data type, the output is able to show more detailed changes. Clearly, the output are mimicking the styles from the input and combining them in a latent way. Furthermore, our system could generate output shape very fast, usually ten to twenty seconds. After preprocessor of the data and pre-train the VAEs. In our experiments, the system is able to finish training of 2000 epochs in less than half of a minute by using a Nvidia Geforce GTX 1080 Graphic Card.

To get higher resolution, this article experiment both on VSL and Point Net. Point Net could provide better results.

## 7 FUTURE WORK

As promised, this article successfully experimented both of the Point Net neural network and VSL, separately, for high-revolution Point Clouds data in the same categories input and for low-revolution voxel data in different categories.

Most of the output data need to convert to mesh files which industries designers and conceptual graphic designers used for their daily design. This means the low revolution voxel data will achieve a smooth surface output in CAD model which can be used in different scenario: physics-based stimulator, scenario-based simulator, and graphic design. Particularly, with the help of stimulator, industries designer can evaluate multiple output based on specific spec technology. For instance, the military aircraft designer will measure speed, acceleration, engine efficiency of different airplane output from the previous experiments based on the CAD model, in order to evaluate different type of airplane such as attach aircraft, reconnaissance aircraft, and fighter aircraft.

The entire training time is between ten to twenty seconds per output, and the pre-trained neural network takes about 1 day to train. This will save huge time and effort for designers to sketch and test the model in different scenario, and implement this system in different industries.

# Appendix

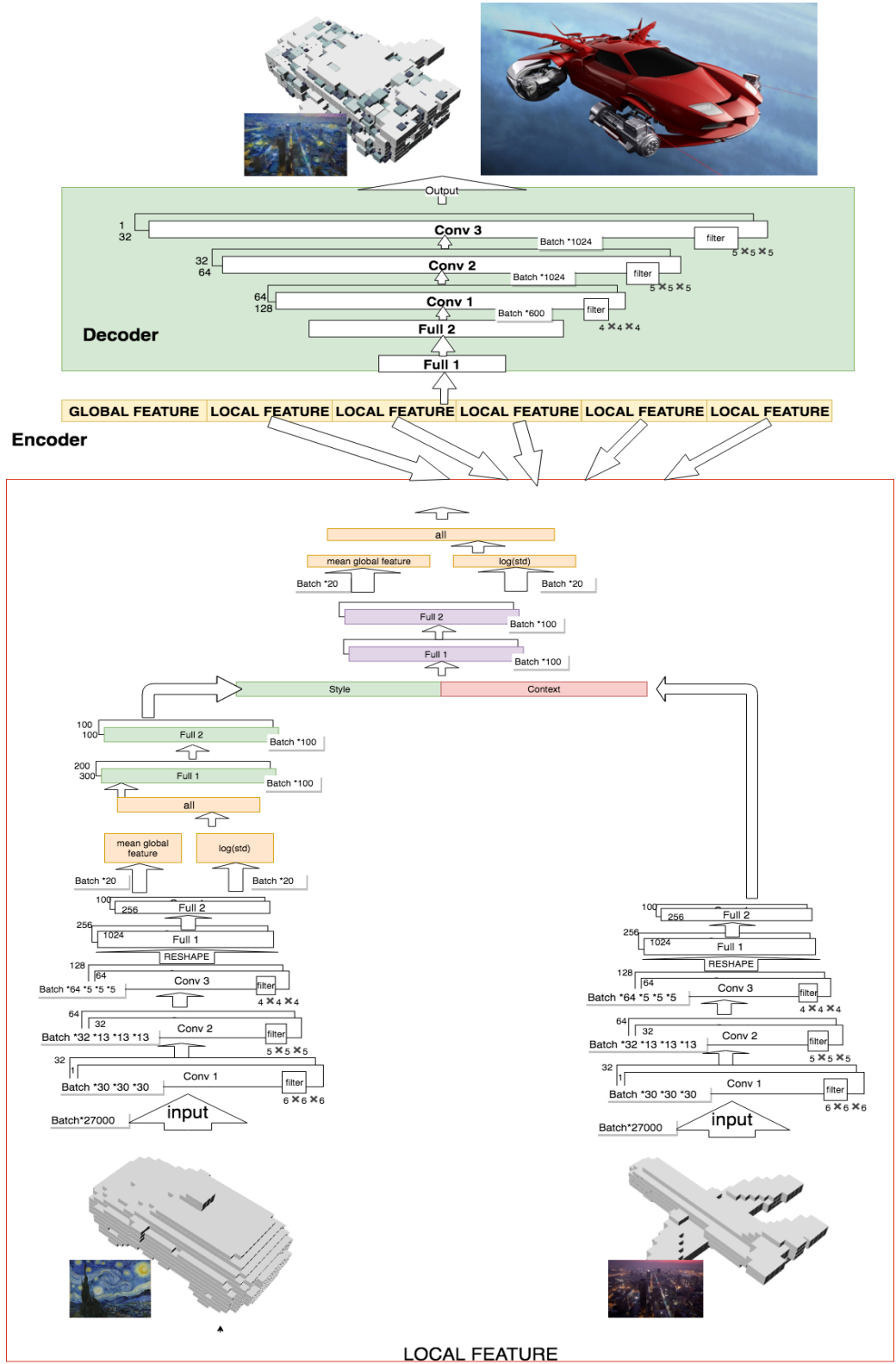

Figure 11: Detail Architecture of Neural Network